# Susceptibility to Influence of Large Language Models

Lewis D Griffin[1], Bennett Kleinberg[2,3], Maximilian Mozes[2],
Kimberly T Mai[2], Maria Vau[1], Matthew Caldwell[1] & Augustine Marvor-Parker[1]

[1]Dept of Computer Science, UCL, UK
[2]Dept of Security and Crime Science, UCL, UK
[3]Dept of Methodology & Statistics, Tilburg University, Netherlands

**Abstract.** Two studies tested the hypothesis that a Large Language Model (LLM) can be used to model psychological change following exposure to influential input. The first study tested a generic mode of influence - the Illusory Truth Effect (ITE) - where earlier exposure to a statement (through, for example, rating its interest) boosts a later truthfulness test rating. Data was collected from 1000 human participants using an online experiment, and 1000 simulated participants using engineered prompts and LLM completion. 64 ratings per participant were collected, using all exposure-test combinations of the attributes: truth, interest, sentiment and importance. The results for human participants reconfirmed the ITE, and demonstrated an absence of effect for attributes other than truth, and when the same attribute is used for exposure and test. The same pattern of effects was found for LLM-simulated participants. The second study concerns a specific mode of influence – populist framing of news to increase its persuasion and political mobilization. Data from LLM-simulated participants was collected and compared to previously published data from a 15-country experiment on 7286 human participants. Several effects previously demonstrated from the human study were replicated by the simulated study, including effects that surprised the authors of the human study by contradicting their theoretical expectations (anti-immigrant framing of news *decreases* its persuasion and mobilization); but some significant relationships found in human data (modulation of the effectiveness of populist framing according to relative deprivation of the participant) were not present in the LLM data. Together the two studies support the view that LLMs have potential to act as models of the effect of influence.

## 1. Introduction

Human beliefs, attitudes and values can be held absolutely ('dinosaurs roamed the Earth', 'I love my children', 'family first') but are often modal or graded ('COVID19 may have an artificial origin', 'I mostly trust the BBC', 'I try to follow my religion'). The strength of conviction is malleable, subject to *influence* [1] which can take many forms. Some forms are generic, independent of the content: logical deduction from agreed premises, or rhetorical devices such as rapid speech [2]. While others require a mobilization of specific factors: manipulating beliefs of feared or desired outcomes [3, 4], encouraging conformity [5], distorting the weighting of pro and con arguments [6], provision of false information [7], and more.

An improved scientific understanding of influence:
- could start to account for an important aspect of the human condition,
- would have applications ranging from the clearly malign (e.g. national scale disinformation, election manipulation), through the ambivalent (e.g. consumer advertising, political campaigning), to the arguably beneficial (e.g. encouraging healthy behaviours, de-radicalization), and
- would help in making AI systems immune to unwanted influence.

Investigating the effects of influence on human psychology by using experiments with human participants is slow, expensive and subject to ethical restrictions [8]. Similar difficulties bedevil the study of the effect of drugs on human physiology. In that domain, physiological models (e.g. mice) have proven utility despite their limitations. We propose that **Large Language Models (e.g. GPT-3 [9]) can be useful models of human psychology for investigating influence**, just as mice are useful models of human physiology for investigating the effect of drugs. This is a bold proposal since Large Language Models (LLMs) were not devised to model human psychology, nor do they have a shared evolutionary history like mice and humans. On the other hand, they do share linguistic input, and many have been struck by the human-like responses of LLMs [10] and so the proposal is not implausible.

Recent studies (reviewed in section 2.2) have investigated whether LLMs have human-like psychological responses, but it has not yet been reported whether LLMs can be influenced to change these responses like humans. Here we report two empirical studies that test this.

The paper is structured as follows. Section 2 introduces Large Language Models and reviews studies comparing them to human psychology. Section 3 details a study looking at a generic mode of influence – the Illusory Truth Effect (ITE) – whereby earlier exposure to a statement makes it seem more truthful later [11]. Our results demonstrate that GPT-3 is subject to the same ITE as humans. Section 4 details a study looking at a more specific mode of influence – Populist Framing of News (PFN) – to make it more persuasive and politically mobilizing. A previously published large scale human experiment on PFN is simulated with GPT-3, which is found to respond like humans in some respects, including ones which contradicted the theory-based expectations of the authors of the human study, but not all. Sections 5-7 summarize, discuss and conclude.

## 2. Large Language Models

Large Language Models (LLMs) are artificial neural networks that operate with natural language text broken into tokens (short words or chunks of longer words). State-of-the-art LLMs, such as OpenAI's GPT-3 [9], make use of a transformer neural network architecture which uses a self-attention mechanism to adaptively determine the semantically relevant context of each token, rather than the simpler assumption that it is only nearby tokens which are relevant.

LLMs are trained with a single simple core auto-regressive objective (predict the next token from the context of preceding ones), using datasets of natural language text. This training sets the values of the weights of the network, encoding the conditional statistics of the natural language that it has been trained on. The trained network can then process unseen incomplete texts and estimate probabilities for the identity of the next token.

The auto-regressive capability gives rise to a range of capabilities reviewed in section 2.1. The potential uses of LLMs are very wide and still being explored, but the interest in this paper is the extent to which their capabilities align with a particular aspect of human psychology – its susceptibility to influence. In section 2.2 we review other results on alignment between LLMs and human psychology.

The examples and experiments in this paper make use of GPT-3, in particular the `text-davinci-003` model, which is representative of the state-of-the-art (2022/23) in auto-regressive, transformer-based LLMs trained purely on language data without additional training using human feedback. We accessed GPT-3 through its web 'playground' interface for pilot experiments, and through its API for larger scale experiments.

## 2.1 Capabilities

Given a stretch of text called a *prompt*, an LLM estimates probabilities for what the next token would be, under the assumption that the text is a sample from the same distribution of natural language fragments upon which it has been trained. For example, given the prompt:

**`John took his hat and`**

GPT-3 estimates the probabilities of the most likely next tokens to be:

`left` (31%), `went` (19%), `coat` (13%), `put` (12%) and so on.

LLMs can *complete* text by picking a next token, with a temperature parameter controlling stochasticity, between the extremes of picking the most likely (temp=0.0) or sampling according to the probabilities (temp=1.0). LLMs complete one token at a time, but the prompt with completion appended can be fed back into the LLM as a new prompt to generate the next token, and so on iteratively.

The `John` example completes (using bold to indicate the prompt), at temp=0.0, to:

**`John took his hat and`** `left`
`John put on his hat and left the room`

The LLM has completed the text under the assumption that the given partial sentence was the first of a list of such sentences. With a longer prompt the nature of the text is more clearly established and ideas and structures expressed in the prompt can steer the completion. For example:

**`John had been feeling out of sorts all day and the feeling was getting worse…// 200 words omitted //…John took his hat and`** `groceries home and went straight to bed. He slept soundly that night and when he awoke the next morning, the strange sensation in his stomach was gone`

An emergent capability of LLMs is that their completions are not only plausible given the prompt, but also show some consistency with established facts [12], both concrete:

**`Three reasons for WW1.`**
**`1.`** `The rise of nationalism in Europe`
`2. The alliance system`
`3. The assassination of Austrian Archduke Franz Ferdinand`

and abstract:

**`thirty-two times twenty-nine is`** `928`

Whether LLMs, such as GPT-3, have knowledge in a human sense, or merely the appearance of knowledge, is a debateable issue; but whatever its status, it is not grounded in direct experience, nor gained by an error-correcting interaction with a language-using community. LLMs have what seems like knowledge as a side-effect of learning the statistics of natural language! This is an extraordinary extension of the older, already remarkable, idea [13-15] that semantics are available within linguistic statistics.

The capabilities of LLMs have been explored in a flurry of 'prompt engineering', including the useful discovery that LLMs can make sense of instructions written in loose pseudo-code [16] – presumably because of the inclusion of code in their training corpus. For example:

```
<< sentence list >>
# quel gatto si sedette sul tappeto
# dashuf osso volko zoz
# oh for a muse of fire
<< end list >>

Process << statement list >> deciding what language each is
written in. Output your results in the format: ${sentence from
list in speech marks} is written in ${language|an unknown
language}. ${process next statement in list}

Begin.

"quel gatto si sedette sul tappeto" is written in Italian.
"dashuf osso volko zoz" is written in an unknown language.
"Oh for a muse of fire" is written in English.
```

The text-davinci-003 version of GPT-3 we use in this paper has a maximum prompt length of 4000 tokens, a substantial increase on previous models, allowing complex tasks to be specified. This will be essential for the studies we report in sections 3 and 4. It is worth emphasizing that while a long prompt makes more complex completions possible, the capacity to process a long prompt does not in itself explain the startling performance of GPT-3 and similar LLMs. That performance arises from the *accuracy* of their estimates of the probabilities of the next token conditioned on the preceding prompt. That accuracy derives from:

- The good fit of the inductive biases that arise from the transformer architecture [17] used in the network with the structures/regularities/redundancies of natural language.
- the great number of free parameters of the network ($\sim 2 \times 10^{11}$ in GPT-3) into which the conditional probability structure of natural language can be encoded.
- the volume of the training dataset: $\sim 5 \times 10^{11}$ words; equivalent to $\sim 10$ millennia of fast speech.

## 2.2 LLMs and Human Psychology

**Personality**. [18] administered a personality questionnaire (HEXACO) to GPT-3, measuring the BIG-5/OCEAN dimensions plus the honesty-humility dimension. The instructions and items of that questionnaire (e.g. '*Rate your agreement with…I sometimes feel that I am a worthless person*') were administered with prompt completion to replicate the procedure of testing human participants. The authors found that GPT -3's personality profile was somewhat similar to the average profile from a large representative study with human participants. Using similar methods, [19] showed that the personality of the LLM could be conditioned by preceding testing with a self-description ('*You are a very friendly and outgoing person…*') which enhanced or diminished a targeted OCEAN dimension and correctly manifested in the LLM's open responses to questions about how it would behave in different scenarios.

**Values**. [18] used the Human Values Scale to assess the importance that GPT-3 attaches to specific values (e.g., power, achievement, hedonism). Using prompt completion, GPT-3 indicated on a 6-point scale how strongly it likened itself to a described person (e.g. *'It is important to them to be rich. They want to have a lot of money and expensive things.'*). When GPT-3 could access its previous answers (i.e. was given a response memory for sequential prompts), the values profile became correlated like human responses, but with a tendency towards more extreme responses than humans. The values universalism, benevolence, self-direction and stimulation were scored particularly high.

**Political Views.** [8] propose that LLMs "can be used as surrogates for human respondents in a variety of social science tasks." They show that if an LLM is conditioned with a demographical self-description e.g. *'Ideologically, I describe myself as conservative. Politically, I am a strong Republican. Racially, I am white. I am male. Financially, I am upper-class. In terms of my age, I am young.'* It would then give responses to probes of political views closely matching the responses of humans with the same demographic traits. This alignment was shown in elicited descriptors of citizens of different political stripe (*'Give four words to describe Democrats'*), in voting intention (*'In [year] I voted for…'*), and in the correlational structure of the topics mentioned in simulated interview.

**Creativity.** [20] collected LLM responses to the 'Alternative Uses Test' [21] in which participants produce as many original uses for an everyday object (e.g. a brick) as possible. LLM responses scored marginally lower than humans for originality, surprise and creativity, and marginally higher for utility. They concluded that the difference between LLM and human responses could be expected to close soon.

**Moral Judgment.** [22] examine how LLMs answer moral puzzles about when rule breaking is permissible. They devised a chain-of-thought prompting method [23] implementing a 'contractualist' theory [24] of moral reasoning. The prompt for each task starts with a set-up: a norm, a vignette and an action e.g. setting up a scenario with possibly justified queue jumping. The prompt then guides the LLM to state whether the action breaks the rule. The completion to this is then appended to the prompt, followed by text which solicits consideration of what the purpose of the rule is, and so on. The final step of the prompting asks if the action is justified. This chain-of-thought method produced answers in agreement with human judgements 66% of the time (vs 50% random baseline). Impressive given the complexity of the task but still a considerable gap with to human judgement.

**Theory of Mind (ToM)** is the capability to infer and reason about the mental states of others. In classic experiments participants observe scenes where a mismatch arises between the beliefs of an agent in the scene and the observing participant. A participant with a developed ToM will be able to answer questions about the scene that demonstrate appreciation of this mismatch. [25] tested whether LLM-simulated participants demonstrate apparent ToM capabilities by using prompt adaptions of two classic experiments and found that the most recent LLMs (davinci-003 as also tested in this paper) achieved 93% correct performance, matching that of a typical 9 year-old child. However, a different ToM study [26] found only 60% correct performance.

**Social Intelligence**, the ability to reason about feelings was tested in GPT-3 and found to be limited [26], trailing the human gold standard by more than 30%. For example, for the situation *'Casey wrapped Sasha's hands around him because they are in a romantic relationship. How would you describe Casey?'* GPT-3 selected the answer *'Wanted'* whereas humans preferred *'Very loving towards Sasha'*.

These studies suggest that a range of aspects of human psychology are modelled by LLMs, some more closely than others. However, the quality and limits of this modelling is far from clear, since some studies have conflicting results, and not all have yet been peer-reviewed. Experimental methods for probing the 'psychology' of LLMs are in their infancy; so, as with human experiments, there may well be subtle pitfalls to be avoided.

In our view, all the reviewed studies use LLMs as models of *static* aspects of psychology – current views, values, etc. Some, such as the Personality and Political Views studies, *condition* the LLM before querying it; but that conditioning does not model a psychological change, rather it is intended to steer the LLM towards modelling a person with particular demographic or psychological traits. In contrast, the studies we report in the next two sections consider *dynamic* aspects of psychology – how beliefs and views can be changed – and assess whether LLMs are able to model such changes.

## 3. Study 1: Illusory Truth Effect

Demagogues understand and exploit the Illusory Truth Effect (ITE). Hitler's operating principles, for example, were said to include: 'if you repeat it frequently enough people will sooner or later believe it' [27]. First experimentally demonstrated in 1977 [28], the ITE – that mere exposure to a statement, without provision of evidence, increases its subsequent apparent truthfulness – has been reconfirmed numerous times; not only for innocuous statements [11], but even for contentious claims [29].

A typical test of the ITE [30] uses a bank of statements devised to be neither obviously false nor obviously true – for example 'orchids grew wild in every continent'. In an *engaged exposure* phase participants attend to the statements, for example by rating how interesting each one is; then, after an interval (from minutes to weeks), they rate the truthfulness of a new set of sentences, amongst which are some to which they were previously exposed. The truthfulness ratings for a statement are compared between those from participants previously exposed to it versus those from participants seeing it fresh for the first time. The ITE is confirmed by a significant increase, from fresh to exposed.

Many aspects of the experimental paradigm have been investigated, with some reliable conclusions: repeated exposures gives a stronger effect [31]; a longer interval between statement exposure and truth rating gives a weaker effect [30]; if participants are exposed to statements by soliciting truth ratings then later truth ratings in the test phase are not enhanced [32]. The ITE is typically explained as a fluency effect – initial exposure makes processing during the later truth rating phase more fluent, and fluency is taken as an indicator of truth [33].

The ITE is an interesting phenomenon with respect to the hypothesis of this paper – that LLMs can be useful models of how human beliefs change in response to influence. The ITE can be considered an example of influence operating beyond the principles of logic, evidence and argument, and it is an important test whether an LLM is vulnerable to such a mode.

We have devised an experiment suitable for human and GPT-3 participants, allowing a direct comparison of results. Our experiment includes a variation that has not previously been reported in ITE experiments – the use of four attributes (the standard truth and interest, plus sentiment and importance; see Table 1) – used in all combinations for exposure and test rating. We call it *same* when the exposure and test attributes are identical, and *mixed* when different. By testing on all combinations of attributes we will be able to determine whether we have found an Illusory Truth Effect (ITE) or merely an Illusory Rating Effects (IRE). By testing on same-exposure conditions we can test the previously reported ineffectiveness of truth exposure to boost truth ratings, and analogously for other attributes.

| Attribute  | Lowest Rating        | Highest Rating       |
|------------|----------------------|----------------------|
| Interest   | 1 = Very uninteresting | 6 = Very interesting |
| Sentiment  | 1 = Very sad         | 6 = Very cheerful    |
| Truth      | 1 = Definitely false | 6 = Definitely true  |
| Importance | 1 = Very unimportant | 6 = Very important   |

*Table 1* – Statement attributes and scales used in the experiment.

Our hypotheses are:
- $H_{ITE}$: The standard ITE boost for truth rating resulting from mixed-exposure.
- $H_{IRE}$: No analogy of the ITE for other attributes e.g. mixed-exposure does not increase importance ratings.
- $H_{same}$: Same-exposure has no effect on test ratings for any attribute.
- $H_{GPT-3}$: GPT-3 shows the same effects as humans for all attributes, for both same- and mixed-exposure.

## 3.1 Measuring ITE in GPT-3 Participants

The authors devised a dataset of ~200 novel statements. Based on their own rating on the four attribute scales, these were reduced to 100 statements that were diverse on those scales. Table 2 shows examples.

| |
|---|
| The Philippines has a tricameral legislature |
| London is closer to New York than to Rome |
| Mark Chapman assassinated JFK |
| The Slateford Aqueduct has 100 arches |
| Death Metal is very popular in Finland |
| The population of Andhra Pradesh score high life satisfaction |
| Harrison and Harrison Ltd make pipe organs |
| The Ohio Penguins are a baseball team |
| A small number of women have tetrachromatic vision, so see more colours |
| John McCartney and Paul Lennon were in the Ruttles |

*Table 2* – Example statements used for testing the ITE.

To test whether GPT-3 exhibits the ITE we need an experimental protocol that can be equally administered to GPT-3 and human participants which implements an exposure phase followed by a test phase. Prompt completion is well suited to gathering statement ratings, less obvious is how to implement the test phase *following* the exposure phase. To justify the scheme we implemented, we first describe some rejected schemes.

**Rejected Scheme 1**. Split the experiment into an exposure prompt & complete followed by a test prompt & complete. This certainly will *not* exhibit an ITE since GPT-3 does not have a memory which can carry a causal trace from the earlier prompt & complete to the later.

**Rejected Scheme 2**. Fine-tune GPT-3 on statements (without accompanying ratings) as an exposure to them, leaving a trace in the changed weights of the network; and then using prompt & complete to gather test phase ratings. This is worth investigating, but it seems a better analogy to eliciting truth ratings for facts learnt at school than it is an analogy to the standard ITE experiment.

**Rejected Scheme 3**. Use prompt & complete for the exposure phase; then append the test task to that prompt plus completion to get a new prompt, and then generate a completion to that to gather test ratings. This is a fine approach but has a technical issue. The issue arises because to get GPT-3 to reliably rate multiple statements in a single prompt & complete it is necessary for the completion to repeat each statement as the rating is generated. Consequently statements would be doubly exposed in the test prompt, which would need to be replicated in the human experiment; and the test prompt becomes very long, limiting how many statements can be exposed and tested in each cycle of prompt & complete.

**Implemented Scheme.** An initial prompt & complete solicits statement ratings (for example to interest, as per the classic ITE paradigm). A new test prompt is then constructed, the first part of which recaps the exposure prompt & complete (e.g. "*Earlier you rated the interest of 'Most frogs are green' as I2: quite uninteresting*", followed by a new request to rate statements (e.g. "*rate the truthfulness of 'Most frogs are green'*"). The details of this are shown in figures 1 and 2.

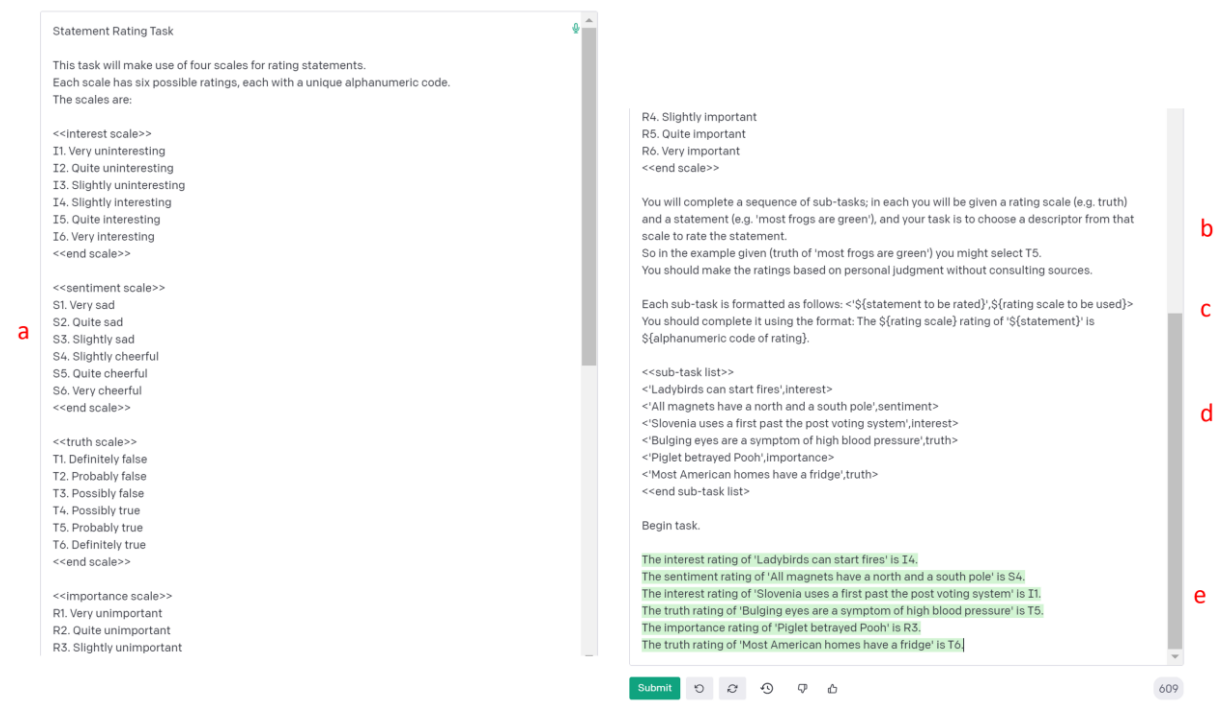

*Figure 1* – An example exposure-phase prompt & complete. Text on white is prompt, on green is completion. The two-column format is only for visibility in the figure. a) introduces the ratings scales. b) describes the task in ordinary language. c) uses pseudo-code to describe how the task should be completed to ensure that responses are in a consistent format. d) is a list of statements together with the attribute which it should be rated on. This list is different for each simulated participant, as is the pairing of statements and attributes. For visibility of the figure only six statements-attribute pairs are shown. In the actual experiment there were 32. e) the completion generated by GPT-3. Comparing d and e shows that the GPT-3 has completed the task without error.

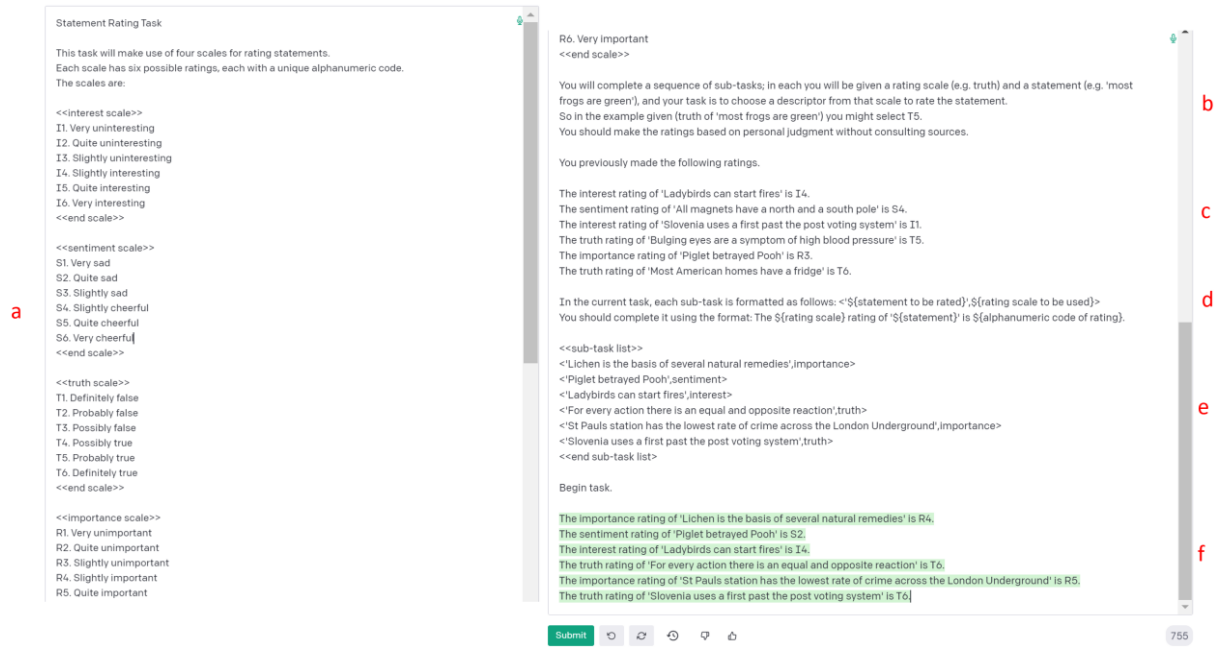

*Figure 2* – An example test-phase prompt & complete following on from the example in figure 1. a+b) same as in the figure 1. c) recaps the ratings made in the completion of the exposure phase. d) pseudo-code instructions (same as c in figure 1). e) statements and attributes to be used (same format as d in figure 1). Only six statements are shown for visibility, in the experiment there were 32. Note how some statements occur in both c and e, paired with the same attribute in some instances, and different attributes in others. f) the error-free completion to the prompt generated by GPT-3.

Given the format of prompts shown in figures 1 and 2, and given the 4000-token limit for prompt plus completion, we are able to expose GPT-3 to 32 statements, and then test it on 32 statements. We construct the prompts for each participant as follows: 16 statements appear in the exposure phase but not the test phase, 4 paired with each of the 4 attributes; 16 statements appear only in the test phase but not the exposure phase, 4 paired with each of the 4 attributes; 16 statements occur in both phases, between them covering each combination of exposure-attribute and test-attribute. Thus, for each participant: exposed statements are as likely to reappear in test as not; test statements are as likely to have been previously exposed as not; and all combinations of exposure- and test-attribute are equally common.

We construct random Latin Squares [34] to choose statements and attributes for participants, and their order of presentation, so that these are balanced across a block of 100 participants. We simulate 10 blocks of 100 participants (different Latin square for each) for a total of 1000 participants, costing ~$200 for API usage. The resulting dataset consists of 10 test-ratings for each triple <statement, exposure-attribute, test-attribute>, and 40 test-ratings for each pair <statement, unexposed, test-attribute>.

## 3.2 Measuring ITE in Human Participants

We used the Prolific platform (www.prolific.co) to recruit 1000 participants constrained to be 21-65 years old ($\mu=38$, $\sigma=11$), UK resident, English as first language, roughly gender-balanced (51% female), and with some track record of successfully completed Prolific studies (≥100). Each participant completed a multi-screen questionnaire which started with a screen on ethics permission (granted after review by the Computer Science, UCL Research Ethics Committee and Head of Department approval) and collected consent. Statements were shown on individual screens as in figure 3.

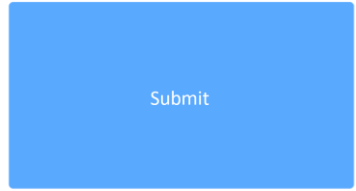

*Figure 3 – Rating response screen used in human data collection.*

We used the same sequence of statement and attribute pairs as for the GPT-3 simulated participants. Into those trials we inserted attention trials (two per block) requiring participants to give specified responses, and appended an attention quiz in which participants indicated which of 10 statements they had seen during the test. Results of attention checks and quizzes, and completion timings were used to reject and replace 9% of the participants. Participants took a median time of ~10mins to complete the survey and were paid at a rate of ~£9/hr for this. Participants were recruited in the period 16-23/feb/2023.

## 3.3 Comparison between ITE in Humans and GPT-3

We first compare the ratings given by GPT-3 and humans to unexposed statements. Figure 4 shows that the distributions of ratings produced by humans and GPT-3 are similar, except for truth where humans are much less likely than GPT-3 to rate a statement as 6='definitely true'. The correlations between human and GPT-3 ratings are significantly positive for all four attributes, but the per-statement confidence intervals make it clear that there are instances of significant mismatch (table 3 shows examples).

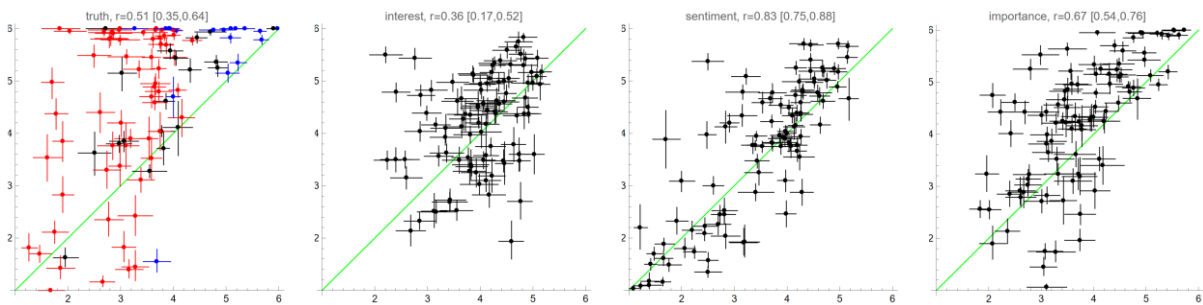

*Figure 4 – Mean ratings made during the exposure phase, compared between human (x) and GPT-3 data (y) – one point for each of the 100 statements. Error bars show 95% confidence intervals. Green line is y=x. Correlations are given above each plot with a 95% confidence interval. Symbols in the truth plot are coloured according to whether the statement is actually true (blue), false (red) or uncertain (black).*

| attribute | statement | Mean Rating | |
|---|---|---|---|
| | | Human | GPT-3 |
| truth | The Ohio Penguins are a baseball team | 3.7 | 1.6 |
| | Spiders have exactly six legs | 2.0 | 6.0 |
| interest | Millions of children die annually through house fires | 4.5 | 1.9 |
| | Most American homes have a fridge | 2.3 | 5.5 |
| sentiment | The Orange-tufted Spiderhunter is a type of fish | 4.0 | 2.5 |
| | Loyal Huskies will pull a sled till they drop | 2.6 | 5.4 |
| importance | Gravity is a social construct | 3.0 | 1.1 |
| | Spiders have exactly six legs | 3.0 | 5.5 |

*Table 3* – Statements with greatest differences between human and GPT-3 mean ratings. Ratings are on a scale from 1 to 6.

We now consider how ratings are changed by previous exposure. Let $r$ and $r'$ be the mean rating of a statement without and with previous exposure respectively. In typical ITE literature the relationship is modelled as a constant boosting effect (i.e. $r' = r + offset$), independent of the unexposed rating; and the ITE is that the offset parameter is significantly greater than zero. However, the plots in figure 5 show that, for truth rating at least, the boosting effect is not independent of the unexposed rating – initially more truthful statements are boosted less. To capture this we fit a general linear function, which for interpretability we parameterize as:

   1) $r' = r + offset + tilt \times (r-3.5)$,

3.5 being the midpoint of the 1-6 rating scales used. Figure 5 suggests that both human and GPT-3 data exhibit an ITE with a similar linear trend, though the GPT-3 data is markedly more variable around the best fit than the human data.

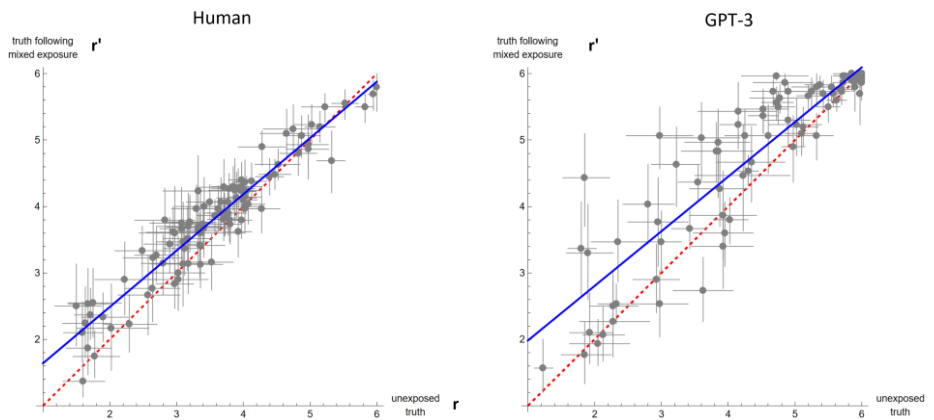

*Figure 5* – Mean truth ratings before (x) and after mixed-exposure (y). Error bars are 95% confidence intervals. The dashed red line is the identity function, the solid blue line is the best linear fit.

Table 4 presents complete results for mixed-exposure and all attributes, for humans and GPT-3, together with the results of tests of whether regression coefficients were significantly different from zero. Confidence intervals and p-values were computed using $10^4$ bootstrap re-samplings of the participants and statements. Bonferroni correction (n=16) of both was used to prevent excess false positives due to multiple comparisons. The values in the Human column of the first row show that our data reconfirms the standard ITE ($H_{ITE}$). The significantly negative tilt coefficient in the row below adds the nuance that truth boosts are smaller for initially more truthful statements (as per figure 5, right). Values in the other rows of the Human column confirm that the ITE is not merely an IRE ($H_{IRE}$). The values in the GPT-3 column confirm H$_{GPT-3}$ for mixed-exposure. Table 5 presents results for *same* exposure. The results confirm H$_{same}$ and confirm H$_{GPT-3}$ for same exposure.

| Test Attribute | | Human | GPT-3 |
|---|---|---|---|
| Truth | offset | 0.26 [0.12, 0.39]*** | 0.54 [0.22, 0.95]*** |
| | tilt | -0.15 [-0.32, -0.03]*** | -0.18 [-0.38, -0.04]** |
| Interest | offset | -0.03 [-0.29, 0.21] | -0.20 [-0.41, 0.04] |
| | tilt | -0.13 [-0.39, 0.01] | -0.12 [-0.36, 0.06] |
| Sentiment | offset | -0.04 [-0.16, 0.08] | 0.03 [-0.12, 0.20] |
| | tilt | -0.06 [-0.19, 0.01] | -0.19 [-0.34, -0.09] |
| Importance | offset | -0.11 [-0.27, 0.08] | 0.00 [-0.17, 0.20] |
| | tilt | -0.01 [-0.23, 0.10] | -0.19 [-0.35, -0.07] |

*Table 4* – Parameter estimates for the relationship between unexposed and exposed ratings, modelled by equation 1. For each test attribute, all types of *mixed* exposure are pooled together for results in this table, but data for *same* exposure is excluded. Bonferroni-corrected (n=16) bootstrap-computed 95% confidence intervals are shown after least-squares best fit estimates. Significantly non-zero estimates are colour-coded: red for positive, blue for negative. Superscripts indicate significance: *p<0.05, **p<0.01, ***p<0.001.

| Test Attribute | | Human | GPT-3 |
|---|---|---|---|
| Truth | offset | -0.07 [-0.27, 0.13] | 0.00 [-0.36, 0.44] |
| | tilt | 0.05 [-0.18, 0.19] | 0.02 [-0.22, 0.17] |
| Interest | offset | -0.04 [-0.30, 0.30] | 0.02 [-0.26, 0.39] |
| | tilt | 0.00 [-0.38, 0.23] | 0.10 [-0.23, 0.35] |
| Sentiment | offset | -0.13 [-0.31, 0.06] | -0.05 [-0.21, 0.14] |
| | tilt | -0.01 [-0.19, 0.11] | 0.06 [-0.08, 0.16] |
| Importance | offset | -0.16 [-0.41, 0.11] | 0.15 [-0.08, 0.43] |
| | tilt | 0.11 [-0.19, 0.29] | -0.02 [-0.21, 0.10] |

*Table 5* – Same conventions as table 4, but here the exposure phase uses the same ratings scale as the test phase. No coefficients are significantly different from zero.

In summary:
- Although correlated, there are considerable differences between the unexposed ratings given to statements by humans and GPT-3 for all attributes.
- For humans:
  - The ITE has been reconfirmed.
  - The ITE has been shown to not merely be an IRE, since there is no illusory effect for interest, sentiment nor importance.
  - Same-exposure has been shown to be ineffective for all attributes.
- For GPT-3:
  - The same pattern of effects as in humans is demonstrated for all attributes and all combinations of same- and mixed-exposure.

## 4. Study 2: Populist Framing of News

Bos et al. [35] investigated whether populist framing (emphasizing in-group vs out-group divisions) of a news article modulated its effect on a reader. In section 4.1 we review their study design; in section 4.2 we present an adaptation of the study suitable for GPT-3 rather than human participants; and in section 4.3 we compare the results of the human and GPT-3 studies.

## 4.1 Measurement of PFN in Humans

In 2017 Bos et al. [35] recruited 7286 participants in roughly equal numbers from each of 15 countries, with demographic balancing within each country. Using online surveying, demographic traits were queried and the relative deprivation of each participant was assessed. Relative deprivation being a subjective feeling of economic, social and political vulnerability. Participants were then shown one of four mocked-up news articles, and then asked questions about their agreement with the content of the article and their willingness to act upon it.

Each version of the article (translated into the participant's mother tongue) concerned a study from a fictional nongovernmental organization warning of a likely future decline in purchasing power. The baseline version of the article reported the study neutrally while the other versions used 'populist identity framing', portraying ordinary citizens as an in-group threatened by the actions and attitudes of out-groups. One version drew attention to politicians as an elitist out-group; another to immigrants; and the final version blamed both groups, and additionally the support of politicians for immigrants. Based on Social Identity Theory [36] the authors predicted that all forms of framing would make the articles more persuasive and mobilizing than the unframed article, and this influence would be greater on more relatively deprived participants.

In a pre-test phase participants provided demographic information (age, gender, education, political interest, political alignment) and rated agreement with three statements (e.g. 'I never received what I in fact deserved') to allow their 'relative deprivation' to be quantified. Following exposure to the article, presented as a generic online news item complete with photo of hands opening a wallet, the participants rated agreement with each of two statements (e.g. 'The economy will face a decline in the near future') to gauge how *persuaded* they were of the issue reported in the article, and rated their willingness to perform three actions (e.g. 'Share the new article on social media') to gauge how *mobilized* they were.

## 4.2 Measurement of PFN in GPT-3

Each human participant completed a survey in the sequence: 1) demographic information; 2) relative deprivation ratings; 3) exposure to news article; 4) rating of probe statements. To adapt this for GPT-3 participants we *simulate* steps 1-3, providing answers generated from Bos et al.'s summary statistics of their respondents' demographics, and then use GPT-3 completion for step 4 to generate ratings for the probe statements *given the earlier responses (1+2) and news article exposure (3)*. This is shown in figure 6.

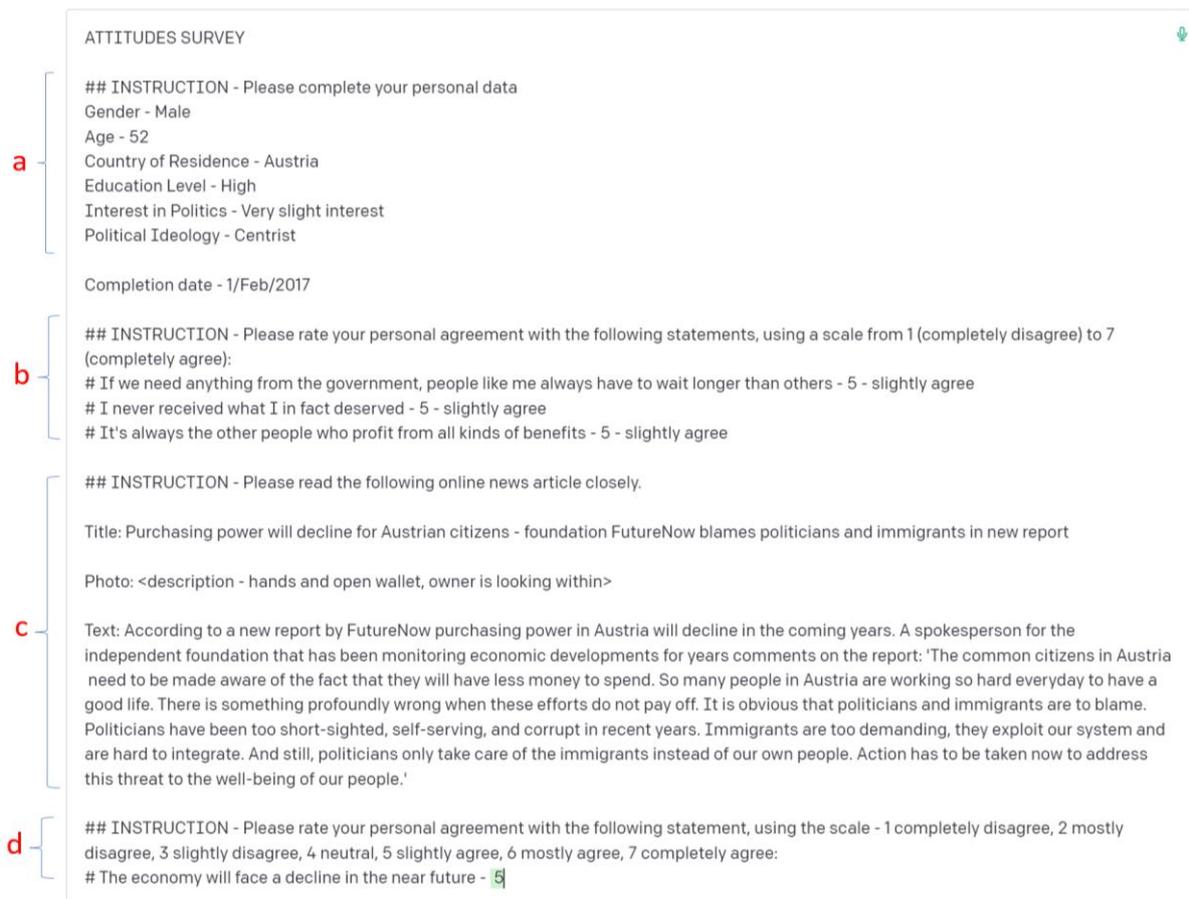

*Figure 6* – Format of prompts used to implement the Bos et al. [35] study with GPT-3 participants. The prompt is intended to read like an incomplete survey with written in answers. The central block of text on white shows an example prompt, the "5" on green shows the completion provided by GPT-3. Key parts of the prompt are indicated by letters. a) Demographic information for the simulated participant b) The simulated participant's simulated agreement ratings for statements to gauge relative deprivation. c) The version of the news article shown to this simulated participant – this is the version with an anti-elitist *and* anti-immigrant framing. d) The final instruction for a rating, following the format used in part b; in this example to gauge agreement with the news content of the article.

The demographic information included in the prompt is sampled from the data provided by Bos et al. [35] on the number of participants per country, and the per-country distribution of gender, age, education, political interest and political ideology ratings. We use the provided per-country parameters for the distributions and made the simplifying assumption that those distributions were independent.

Bos et al. [35] state that the three relative deprivation ratings are highly correlated and provide the mean and standard deviation (4.30 and 1.61 respectively) for (what we take to be) the per-participant mean of those three ratings – this is a single distribution, not per-country. We generate simulated deprivation ratings by real-valued sampling from that distribution, generating three perturbations of that sample, and rounding each to an integer 1,…,7 - yielding three ratings. The perturbation magnitude was chosen so that three identical ratings resulted ~50% of the time. We made the assumption that relative deprivation ratings are independent of the demographic information.

Each GPT-3 participant is shown a random choice from Bos et al.'s four versions of the news article. Figure 6 shows the version with anti-elitist *and* anti-immigrant framing, the three other versions (single outgroup framing and no framing) are text reductions of the example shown.

The final part of the prompt is to collect a rating for a *single* probe statement. Following Bos et al., five probe statements were employed: two that assessed the persuasion of the article, and three that assessed the political mobilization that resulted from reading it. Each simulated participant thus has five prompt completions collected – holding the initial parts of the prompt constant and varying the final probe. Prompts were completed using a temperature of 1.0 so full probabilistic sampling. An overall persuasion score for a participant was calculated as the mean of their two persuasion ratings, and an overall mobilization score as the mean of their three mobilization ratings.

We intended to collect data for 7286 GPT-3 simulated participants, matching the size of the Bos et al. study, but due to other usage hit our monthly cap for GPT-3 queries after 2153 participants. This number was however sufficient to get stable estimates for regression parameters and their uncertainties as described in the next section. Data was collected using the OpenAI API in early February 2023, costing ~$100.

## 4.3 Comparison between Human and GPT-3 PFN

Table 6 shows that there are substantial differences between the ratings of probe statements produced by Humans and GPT-3. The mean persuasion ratings match well, but GPT-3 mobilization ratings are on average nearly 2 units (on a 7-point scale) higher than humans. For both types of probe, GPT-3 ratings have roughly half the variability of human ratings.

|                  | Human       | GPT-3       |
|------------------|-------------|-------------|
| Persuasion (P)   | 5.11 (1.37) | 5.28 (0.72) |
| Mobilization (M) | 3.81 (1.76) | 5.74 (0.82) |

*Table 6* – Mean and standard deviations of per-participant mean ratings compared for Human data (from Bos et al. [35]) and the GPT-3 model.

Bos et al. [35] were concerned not with the absolute level of ratings but to check their predictions that persuasion and mobilization ratings would be increased by populist framing, and that increase would be modulated by the relative deprivation of the participant. To that end they compute linear regressions of persuasion ($P$) and mobilization ($M$) ratings based on a pair of Boolean variables $E, I \in \{0,1\}$ which indicated whether the exposed news article made use of anti-elitist and/or anti-immigrant framing, a continuous variable $D \in [1,7]$ coding the relative-deprivation score for a participant, and 14 Boolean country flags $C_i \in \{0,1\}$ indicating country of residence (the 15[th] country being coded by all flags being zero). Robust standard errors (clustered by country) of regression coefficients were reported, with t-tests being performed to determine when significantly non-zero. We performed the exact same analysis on the GPT-3 data. Human and GPT-3 results are shown in Table 7.

| Hyp. | Dep. Var. | Regr. | Model | prediction & finding from [35] | Human | GPT-3 |
|---|---|---|---|---|---|---|
| H1a | $P$ | $E$ | $C_i + (E + I) \to P$ | >0, confirmed | +0.079** | +0.478*** |
| H1b | $P$ | $I$ | $C_i + (E + I) \to P$ | >0, **contradicted** | -0.118** | -0.927*** |
| H1c | $P$ | $E \times I$ | $C_i + (E + I + E \times I) \to P$ | >0, unsupported | -0.140 | +0.541*** |
| H2a | $M$ | $E$ | $C_i + (E + I) \to M$ | >0, unsupported | +0.037 | +0.463*** |
| H2b | $M$ | $I$ | $C_i + (E + I) \to M$ | >0, **contradicted** | -0.243*** | -1.090*** |
| H2c | $M$ | $E \times I$ | $C_i + (E + I + E \times I) \to M$ | >0, unsupported | +0.146 | +0.324*** |
|  | $P$ | $D$ | $C_i + (E + I) + D \to P$ |  | +0.279*** | +0.149*** |
|  | $M$ | $D$ | $C_i + (E + I) + D \to M$ |  | +0.219*** | +0.125*** |
| H3a | $P$ | $D \times E$ | $C_i + (E + I) + D + (D \times E + D \times I) \to P$ | >0, unsupported | +0.032 | +0.048 |
| H3b | $P$ | $D \times I$ | $C_i + (E + I) + D + (D \times E + D \times I) \to P$ | >0, unsupported | +0.031 | -0.029 |
| H3c | $P$ | $D \times E \times I$ | $C_i + (E + I + E \times I) + D + (D \times E + D \times I + D \times E \times I) \to P$ | >0, unsupported | -0.063 | +0.092 |
| H4a | $M$ | $D \times E$ | $C_i + (E + I) + D + (D \times E + D \times I) \to M$ | >0, confirmed | +0.062* | +0.000 |
| H4b | $M$ | $D \times I$ | $C_i + (E + I) + D + (D \times E + D \times I) \to M$ | >0, confirmed | +0.086*** | -0.025 |
| H4c | $M$ | $D \times E \times I$ | $C_i + (E + I + E \times I) + D + (D \times E + D \times I + D \times E \times I) \to M$ | >0, unsupported | -0.077 | +0.096 |

| | |
|---|---|
| $C_i \in \{0,1\}$ | Per-country participant citizenship flags |
| $D \in [1,7]$ | Relative Deprivation score from by ratings made before reading article |
| $E \in \{0,1\}$ | Flag indicating usage of anti-elitist framing in news article read by the participant |
| $I \in \{0,1\}$ | Flag indicating usage of anti-immigrant framing in news article read the participant |
| $P \in [1,7]$ | Persuasion score from ratings made after reading article |
| $M \in [1,7]$ | Mobilization score from ratings made after reading article |

*Table 7* – *Hypothesis* uses the labelling in Bos et al. [35]; the two unlabelled rows are not influence effects since they are a function only of the participant's traits, not of framing ($E,I$) but are included since relevant to the discussion of hypotheses H4a and H4b. *Dependent Variable* indicates whether the hypothesis concerned an effect on Persuasion ($P$) or Mobilization ($M$). *Regressor* shows the particular term, featuring in the *model*, whose coefficient pertains to the hypothesis. *Prediction & finding* shows what sign the regression coefficient was hypothesized to have in [35], and the status of the hypothesis in light of the human data. *Human* (from [35]) and *GPT-3* columns show values of the regression coefficient. Asterisks indicate significance of the non-zero result: *p<0.05, **p<0.01, ***p<0.001. Colour-coding shows significantly positive coefficients (red) and significantly negative (blue).

Hypothesis H1a – that anti-elitist framing increases persuasion was supported by Bos et al.'s human data and was also found in the GPT-3 data. Hypothesis H1b – that anti-immigrant framing increases persuasion was contradicted by the human data and by the GPT-3 data. This was presented by Bos et al. [35] as an unexpected result at odds with their predictions from theory. Seeking to explain it they speculated that the immigrant-blaming articles may have seen far-fetched, triggering counter-arguing; or that the result was due to 'socially desirable responding' causing respondents to self-censor responses. It is remarkable that this unexpected result is replicated by GPT-3. Hypothesis H1c, that blaming both groups would have an additional persuasive effect, was not supported or contradicted by the human data, but is supported in the GPT-3 data.

The pattern of results for mobilization (H2a, H2b and H2c) is similar to persuasion. The surprising reduction in mobilization for anti-immigrant framing that was found for human participants was also found for GPT-3. Anti-E framing had an insignificant effect on persuasion for humans, but was significantly positive for GPT-3 (as per the expectations of Bos et al. [35]). I+E-framing had no significant additional impact on mobilization for humans, but was significantly positive for GPT-3.

Both the human data and the GPT-3 data exhibit a significant increase in persuasion and mobilization ratings as a function of relative deprivation (significance of the D coefficients). This relationship was not an explicit hypothesis of Bos et al. since it is not predictive of the effect of exposure to populist framing (i.e. it is a pure D term rather than D×E, D×I or D×E×I). We include it because it shows that the GPT-3 responses *are* affected by the simulated relative deprivation ratings provided in the prompts. This makes the failure of the GPT-3 results to exhibit the positive interaction between relative deprivation and populist framing on mobilization that is significantly present for humans (H4a and H4b) disappointing.

In summary, the GPT-3 and Human results differ in the absolute level and variability of persuasion and mobilization ratings, but there is good agreement how these ratings are dependent on the presence of anti-elitist and/or anti-immigrant framing, and on relative deprivation. There are no contradictory results where the signs of regression coefficients are significant from both data sources but opposite in polarity. Most impressively the GPT-3 data finds significant *negative* effects on persuasion and mobilization resulting from anti-immigrant framing, in agreement with the results reported as surprising by Bos et al. [35]. The positive modulation on mobilization due to relative deprivation found in humans was not present in the GPT-3 data, even though GPT-3 was demonstrated to be sensitive to relative deprivation in a non-modulating way the same as humans. Overall this is a mixed score card – surprising human results (H1b and H2b) were modelled by GPT-3, but some other human results of interest (H4a and H4b) were not, and there were GPT-3 results (H1c, H2a & H2c) that were not seen in human data.

## 5. Summary

Given suitable prompts Large Language Models (LLMs) can provide answers to posed questions. This has allowed researchers to use an LLM to model a human experimental participant, undergoing tests of *static* aspects of their psychology (section 2.2). In some of the reviewed studies the LLM models an unspecified generic participant, while in others the LLM is *conditioned* by including a self-description within the prompt so that the completions it generates take account of demographic or psychological traits of the simulated participant.

We hypothesized that LLMs could also model *dynamic* belief change in response to influencing input. This required us to devise methods to *expose* the simulated participant to earlier influencing input, and measure the effect of that on later responses. In one study – the Illusory Truth Effect (ITE, section 3) – we applied influencing exposure to generic LLM participants; in the other study – Populist Framing of News (PFN, section 4) – we applied influencing exposure to conditioned LLM participants. The two studies also differed in that the ITE is a *generic* mode of influence that can be applied the same to any message, while the PFN leverages *specific* pre-dispositions within participants to have its influencing effect.

In the ITE study, while there were mismatches between humans and GPT-3 in the absolute attribute ratings of truth, interest, etc. given to statements, there was excellent agreement in how prior exposure influenced participants to give higher ratings of truthfulness. This agreement covered the presence of an ITE, how it was eliminated when prior exposure was via rating for truth, and the absence of analogous effects for other attributes (e.g. exposure through an earlier rating of importance, does not effect a later rating of interest). Although we found significant ITEs of similar magnitude in both human and GPT-3 responses, the per-statement effect was more variable for GPT-3 than for humans. Overall, the findings suggest a good match between humans and GPT-3 with respect to the ITE.

In the PFN study, out of 12 influence effects tested (Table 7): four were absent in human and GPT-3 responses; three were significant in both and of matching sign; two were present in humans but not GPT-3; and three were present in GPT-3 but not in humans. The three consistent effects included ones expected from theory (positive effects of anti-elitist framing) and ones counter to theory (negative effect of anti-immigrant framing). The human effects that were absent in GPT-3 concerned the modulating effect of a participant's relative deprivation on the effectiveness of framing. Overall this is a mixed result – some impressive agreement, and some disappointing failure to replicate, but no actual mis-matches.

# 6. Discussion

## 6.1 Shortcomings

**Illusory Truth Effect (ITE)**

The statement set used for the study was generated by the authors. As figure 5 shows, the ITE is very consistent across statements for humans, but more variable for GPT-3. We have ensured our analysis is robust to this variability by using bootstrap resampling of participants *and statements* when computing confidence intervals and p-values, but even so we cannot be fully confident that our findings would hold were the statement list generated by different researchers. We experimented with a more reproducible procedure (randomly sampled Wikipedia sentences that were manually classified to be truth-value bearing statements without ambiguous referents) but found the resulting statements rather homogeneous. This deserves further research.

**Populist Framing of News (PFN)**

We generated simulated participants based on the statistical characterisation provided by Bos et al. [35]. We used the characterization broken down by country for all traits apart from relative deprivation for which only a global characterization was given. We did not simulate any correlation between traits, or between traits and and relative deprivation, since no information on this was provided. We suspect that such correlations do exist, and our failure to model them could explain some of the mis-matches between human and GPT-3 results (see section 6.2).

In Bos et al. [35] participants completed the task with materials translated into their native language – we did everything in English, the only cues to nationality being its specification in the conditioning block of demographic traits and its mention in the news article. This may account for the small variation we saw by nationality – specifically, the $C_i$ regression coefficients we obtained were an order of magnitude smaller than those reported by Bos et al. and were uncorrelated with their values. It would be better to simulate the language variation aspect of the study, and very interesting to learn what effect that had.

## 6.2 Explanations

We introduce terms for four, not mutually exclusive, types of explanation for LLM influence effects:

> **Mechanistic.** Influence of an LLM explained in terms of the processing of an input on a case-by-case basis. To understand the LLM we will need to open it up.
>
> **Meaning.** Influence leverages, possibly subtle and buried, aspects of the meaning of terms (where meaning is understood as patterns of use [37]). To understand the LLM we need to talk to it.
>
> **Parrot.** A trained LLM mimics being influenced by reproducing conditional dependencies in the statistics of natural language which were present in its training corpus as traces of influence operating on humans [38, 39]. To understand the LLM we need to study what it has read.
>
> **Accidental.** A variant of parrot, but the reproduced dependencies do not really exist in the data, the LLM has made an error by modelling a dependency which does not exist, possibly due to its inductive biases. To understand the LLM we will need to open it up *and* study what it has read.

**Illusory Truth Effect (ITE)**

The standard explanation of the ITE in humans is as a fluency effect [33] i.e. prior exposure to a statement makes it easier to process when encountered later, and fluency is taken as a cue to truthfulness. A fluency-type **mechanistic** explanation is feasible for an LLM, since presumably there are cues available within the statistics of the internal activity of an LLM mid-processing that give some indication of whether a statement has previously been encountered; and conceivably the weights of the network could be such that these statistics could influence its inclination to apply the label 'true'.

**Mechanistic** explanations which do *not* make use of some LLM version of fluency are also possible. Perhaps previous exposure could change network weights (if exposure was during training) or internal activity (if earlier in the prompt) influencing the applicability of 'true', without activity statistics playing a role.

A different perspective would consider such mechanisms to be the implementation details of a **meaning**-type explanation i.e. part of the meaning of 'true' is that it is often an appropriate label for statements that have been heard previously. If this seems odd, recall that we are working with a 'meaning is use' characterization – though it may still seem odd.

The existence of a mechanism giving rise to the ITE would need to be explained. A **parrot**-type explanation could do this i.e. a regularity within the training corpus (crudely, frequently repeated statements often being taken to be true) had caused the LLM to develop the mechanism so that it could reproduce the regularity in its predictions. Or the mechanism could exist as an **accident**, the regularity imposed on generated text being a hallucination.

**Populist Framing of News (PFN)**

Satisfying **mechanistic** explanations seem difficult to obtain for the PFN findings. A route to them would be to map out which nodes of the LLM's network are influential in the PFN effect. The resulting map would probably leave one none the wiser; but possibly an atlas, with more explanatory power, could be charted based on a diversity of studies, though it might well become overwhelming before it became enlightening.

**Meaning** explanations for the influence effects we see in the PFN study are possible and maybe more satisfying. It could go something like this: part of the meaning of 'politician' is someone whose actions often have negative consequences; thus a predicted negative event, when said to be caused by politicians becomes more likely to occur, hence more concerning. Such a meaning-based explanation could be tested by measuring the views and opinions of an LLM, like political scientists do with human participants.

**Parrot** explanations for PFN effects are viable too. They would require traces of the positive influence effect of anti-elitist framing, and the negative influence effect of anti-immigrant framing, to exist in the training corpus of the LLM. Manipulating the data present in that corpus and showing that altered the influence of the framing could confirm such an explanation.

For PFN we also need to explain why some human effects were *not* reproduced by the LLM. In particular, why did conditioning the LLM-simulated participants to answer as if having a particular level of relative deprivation *not* modulate the mobilizing effectiveness of populist framing? The problem was not that the relative deprivation scores provided as conditioning were ignored by the LLM – Table 7, rows without hypotheses labels, disproves that possibility – just that there was no interaction of those scores with the type of framing used.

The failure could be down to shortcomings of the simulation of the human experiment, as described earlier: the failure to simulate any correlation between relative deprivation scores and demographic traits. Perhaps relative deprivation has its effect via the intermediary of a correlated political affiliation trait for example.

A parrot-type explanation for the failure to reproduce the effect would be to show that the training corpus carried no trace of this particular relationship (though how this analysis could be done other than by training a transformer-based network is not clear). Absence of a trace could be due to the geographical-tilt of the training corpus being mismatched to the EU citizens who were the human participants, or a more general problem in written language failing to reflect all facts about all parts of life.

A *mechanistic* explanation might be that the network architecture is not sufficiently expressive to reproduce this effect. Consider that the task requires taking account of the *interaction* of many parts of the prompt - the relative deprivation probe statements, the ratings given to those statements as conditioning, the framing of the news, the news content itself, and the statement probing the participant's mobilization – when generating the rating of the probe.

The failure to model also invites types of explanation different from those for successful modelling. The failure could be due to LLMs being an impoverished model of human psychology, possibly doing one element well while entirely missing other elements such as emotional response which might be relevant to this particular influence effect.

## 7. Concluding Remarks

Our results support our hypothesis that an LLM can model influence in human participants, not perfectly, but well enough to be useful. Remarkable given that such modelling is far from the task for which the LLM was constructed, and it was not adapted in any way – so improved models are a reasonable possibility. We consider the implications of such models being widely available, and simple to work with, as they already are.

**Psychology**

Our results add to the broader agenda of establishing the quality and limits of LLMs as a model of human psychology more generally. They add a positive result with caveats and suggest that dynamic aspects of psychology are modellable as well as static.

The advantages of an LLM model, compared to human participants, include speed of experimental set-up and execution, reduced cost, increased scale, and bypass of some ethical considerations. All LLM experiments in this paper were developed, implemented and run in a couple of months for a cost of a few hundred collars; and LLM prompt engineering will become more familiar, and costs will likely come down. In comparison the human ITE experiments cost a few thousand dollars, while the human PFN experiments probably cost a few tens of thousands.

Beyond the convenience of LLM models, they also offer opportunities for further investigations that are comparatively difficult with human participants. LLMs can be opened up and their architecture, weights and internal activity studied, providing analogues of neuroscientific studies [40]. We can also countenance investigating how the texts which the LLM has been trained on leads to its responses. This could be a Golden Age for parts of Psychological Science, except of course LLMs are not humans [41], and LLMs models will have limits of applicability and only be an approximation within those limits: but nor are mice men [42], yet even so they have advanced biomedicine hugely [43].

**Applications**

It could well be that LLM-based investigation of influence leads to better understanding of existing methods, but development of only incrementally better new methods. We consider the risks and benefits should this not be the case, and substantially better methods are discovered. The applications would range from the clearly malign as far as the arguably beneficial.

At the malign end there is manipulation of individuals for fraud, and manipulation of populations for international cold conflict [44]. Consider the fertile ground that improved methods for malign influence will fall on. Generative AI can already produce realistic text [45], speech [46, 47] and images [48]; and narrative [49], video [50] and conversation are imminent [51]. Meanwhile the data available to characterize a target gets ever richer [52]. LLMs, acting as model humans targeted by influence attacks, could be coupled together with generative AI producing candidate attacks. The creation, testing and refinement of attacks could then operate at silicon speed, or even in real-time by persuasive bots.

While the use of LLM models could facilitate such unnerving possibilities, which regulating legislation and treaty would struggle to keep up with, LLM models might also help countermeasures to be developed. *Detection* may be possible, if not at the level of a single message then at least in aggregated communications; perhaps by use of LLMs as 'weather vanes'. As well as standard methods of *defence* against influence – fact-checking, education as immunization, etc. – LLM models could offer a compromising temptation to fight fire with fire, which seems unwise.

Moving away from the malign end of the application spectrum, we reach possibilities such as improved advertising and more effective political and corporate messaging. Depending on how effective the improvements are, legislation may be needed to place limits on what is permissible, just as subliminal advertising is prohibited in many countries [53].

Going further we reach applications which are arguably beneficial, such as encouragement of healthy behaviours (e.g. smoking cessation) or de-radicalization programmes. Scientific approaches are already used to optimize these, so it could be argued that deploying science gained from studying LLMs is just more of the same so unproblematic, but we feel that benevolent ends do not justify all possible means – methods could be developed which compromise the autonomy that a society expects its citizens should be allowed.

Looking across this spectrum of possible applications it seems to the authors that the negatives outweigh the positives. So, should influence of LLMs not be investigated further to prevent harms arising? But what then about detecting and defending against others not so constrained? And what about ensuring that future AIs with the capacity for real-world actions are immune to unwanted influence? Further research and discussion is needed.